# FoodReact-AR: Learning Audience Reactions that Zero-Shot LLMs Miss in Arabic Food-Insecurity Discourse

**Wajdi Zaghouani[1], Mabrouka Bessghaier[1], Salma Azouzi[2], Shimaa Amer Ibrahim[1]**
[1]Northwestern University in Qatar, Doha, Qatar
[2]Independent Researcher, Tunisia
{wajdi.zaghouani, mabrouka.bessghaier, shimaa.ibrahim}@northwestern.edu
salmaazouzi220@gmail.com

## Abstract

Facebook reactions record how an audience received a post, a signal absent from the post's text. We treat the six differentiated reactions on 74,241 Arabic food-insecurity posts from public Facebook pages (2020–2024) as compositions, apply a centred log-ratio transformation, and cluster them into four audience profiles: Solidarity (Love-dominant), Dismissal (Haha-dominant), Grief (Sad-dominant), and a flat Ambivalent profile. Dismissal peaks in 2022 (22.3%), the year of the wheat price shock, and Grief in 2024 (14.0%), during the Gaza food emergency; a page-clustered GEE confirms both shifts (odds ratios 2.18 and 1.34), and they survive minimum-reaction filtering, alternative zero replacements and clustering methods, and bootstrap resampling. We then test whether the profiles can be predicted without reactions. A zero-shot Arabic LLM recovers Grief for 83.8% of posts but Dismissal for only 3.4%, and in a matched ablation, showing it the reaction percentages does not help. A simple classifier trained on the reaction-derived labels and tested on unseen pages recovers Dismissal for 60.0% (macro-F1 0.56 against 0.43), mainly from mentions of officials and institutions. Derisive audience responses are thus learnable from text, yet a zero-shot LLM misses them. We release the corpus, LLM and human annotation layers, and an evaluation protocol as FoodReact-AR.

## 1 Introduction

Food insecurity is among the most pressing challenges facing the Middle East and North Africa (MENA). Imports cover around 40% of the region's caloric needs (Le Mouël et al., 2023), and Russia and Ukraine supplied 34% of its grain imports between 2015 and 2020 (Rahimi et al., 2023). Between 2020 and 2024 the region faced three compounding shocks: the COVID-19 pandemic, the 2022 wheat price shock that followed the invasion of Ukraine, and the 2023–2024 war in Gaza, which produced famine-level food insecurity (Hassoun et al., 2024).

Facebook is a primary arena in which Arab publics encounter and respond to this news (Khamis and Dogbatse, 2024). Its six differentiated reactions (Love, Care, Wow, Haha, Sad, Angry) give a structured, language-independent record of collective response (Anwar and Giglietto, 2024). For Arabic NLP, reactions matter for a specific reason: they register how an audience received a post, which may differ from what the post says. A ministry's announcement of a food-security budget can be written in neutral, even celebratory, language and still be met with derisive laughter.

Two obstacles have kept this signal from being used well. First, reactions are compositional: the shares on a post sum to one, so an increase in one reaction mechanically lowers the others. Treating them as independent variables yields spurious correlations (Aitchison, 1986; Greenacre et al., 2023). Second, it is unknown how much of the reaction signal is already recoverable from text, which determines whether reactions add information beyond what a text classifier provides.

We address both with a corpus of 74,241 Arabic food-insecurity posts from public pages, most administered in MENA countries, and ask three questions:

**RQ1:** What audience reaction profiles emerge when reactions are treated as compositions?

**RQ2:** How are these profiles distributed across source types, page countries, and crisis periods?

**RQ3:** Can a post's reaction profile be predicted from its text and source metadata, by a zero-shot LLM and by models trained on reaction-derived labels, and which profiles resist prediction?

Our contributions are:

- A compositional pipeline (CLR transformation followed by clustering) that yields four interpretable audience profiles, a map of their distribution across sources, countries, and crisis periods, and robustness and page-level checks of the central temporal contrast (§4–§5.3).
- Evidence that reaction-derived labels teach what zero-shot LLMs miss: an Arabic LLM recovers the Dismissal profile for 3.4% of posts, and no better when shown the reactions, while a simple page-disjoint supervised baseline recovers it for 60.0% (§5.5, §5.6).
- FoodReact-AR, a research resource with four layers (74,241 clustered posts, 6,225 reaction-blind LLM labels, 800 posts with reaction-informed LLM and independent native-speaker labels, and a 294-post matched ablation) and an evaluation protocol for audience-profile prediction (§6).

## 2 Related Work

**Social media and food insecurity.** Molenaar et al. (2024) applied sentiment analysis and LDA topic modelling to 38,070 food-security tweets and found predominantly positive sentiment even during crises. Goetz et al. (2023) showed that anger, disgust, and fear in COVID-era food-security tweets correlated with contemporaneous food insufficiency rates. A scoping review notes growing interest in computational text analysis for food security, dominated by English-language Twitter data (Meeske et al., 2025). Arabic-language Facebook discourse, the main venue for much of MENA, is largely absent from this work.

**Reactions as an emotional signal.** In NLP, reactions have served as distant supervision for emotion classification (Pool and Nissim, 2016). Negative reactions predict higher engagement, and Angry and Haha cluster around politically contested content (Anwar and Giglietto, 2024); reaction patterns also shape commenting (Masullo, 2022) and differ by crisis framing (Zannat, 2022). Reactions are valid but unstable emotional proxies (El Moudene et al., 2023; Paolillo, 2023). Haha is the clearest case, as it often marks sarcasm or derision, a phenomenon that Arabic sarcasm resources document in text (Abu Farha and Magdy, 2020). Closest to our setting, Freeman et al. (2019) predicted Facebook reaction patterns on posts linking to scholarly articles, and Freeman et al. (2020) measured the diversity of those reactions. Anger is known to drive visibility in political communication, leaving open whether non-negative emotions can do the same (Gerbaudo et al., 2023).

**Compositional data.** Data that carry only relative information, such as proportions, live on a simplex. Log-ratio transformations map them to Euclidean space where standard multivariate methods are valid (Aitchison, 1986; Greenacre et al., 2023; Hu and Satten, 2023). We combine this representation with clustering to profile audience responses in Arabic, and with reaction-blind prediction to test how much of each profile text and metadata can anticipate.

## 3 Data

We collected posts from public Facebook pages through the CrowdTangle API[1] using three Arabic search phrases: الأمن الغذائي (*al-amn al-ghidhāʾī*, 'food security'), أزمة غذائية (*azma ghidhāʾiyya*, 'food crisis'), and إنتاج غذائي (*intāj ghidhāʾī*, 'food production'). The full collection contains 164,503 posts published between July 2014 and July 2024 by 24,515 pages administered from 106 countries. MENA-administered pages account for 92.9% of posts (Egypt 22.2%, Iraq 12.5%, Syria 7.9%, Algeria 6.1%, and 17 other MENA states).

**Filtering.** We removed 27 sponsored posts (164,476 remain) and then applied two criteria: (1) publication on or after 1 March 2020, which aligns the window with the onset of COVID-19 and the spring-2020 rollout of the Care reaction; and (2) at least one differentiated (non-Like) reaction, which the composition requires. The resulting analytical set contains **74,241 posts**.

**Crisis periods.** We divide the window into four periods: COVID-19 (March 2020 to December 2021), Ukraine/wheat (2022), post-crisis (2023), and Gaza food emergency (January to July 2024).[2]

[1]CrowdTangle was Meta's public-content research tool. It was discontinued on 14 August 2024 and replaced by the Meta Content Library.

[2]The first IPC acute food insecurity analysis for Gaza was published in December 2023: https://www.ipcinfo.org/ipc-country-analysis/details-map/en/c/1156749/.

The 2023 period includes October to December 2023, after the war began; monthly trends (Figure 1) show where shifts occur within periods.

**Page country.** CrowdTangle records only the country from which a page is administered. It describes where content is produced, not where the audience is.

# 4 Method

## 4.1 Compositional Representation

We use the six differentiated reactions. Like is excluded because it is the default, low-effort response; users who want to express a specific emotion select one of the other six. For each post we (1) replace zero counts with a pseudocount of 0.5 (Martín-Fernández et al., 2003), (2) convert counts to proportions $x_1, \ldots, x_6$, and (3) apply the CLR transformation:

$$\mathrm{CLR}(x_i) = \log \frac{x_i}{\left(\prod_{j=1}^{6} x_j\right)^{1/6}} \quad (1)$$

A CLR value of zero means that a reaction is at its geometric-mean share; positive values mean over-representation. For example, a post with Love = 80, Care = 10, Wow = 5, Haha = 3, Sad = 1, Angry = 1 has CLR values of +2.82 for Love and −1.57 for Sad and Angry. For strictly positive counts, CLR depends only on ratios, so posts with 100 and 10,000 reactions in the same proportions receive identical coordinates and clustering captures the shape of the response rather than its volume. Zero replacement breaks this invariance for sparse posts: a post with 20 Love and no other reactions has a Love coordinate of 3.07, one with 2,000 Love has 6.91 (§4.2). Because CLR distances equal Aitchison distances, clustering in CLR space is equivalent to clustering in any orthonormal (ILR) coordinates (Aitchison, 1986).

## 4.2 Clustering and Choice of K

We ran *k*-means (scikit-learn; 50 initialisations; seed 42) on the CLR coordinates for $K \in \{2, \ldots, 15\}$ and examined inertia, the silhouette score (Rousseeuw, 1987), and the Calinski–Harabasz index (Caliński and Harabasz, 1974) (Appendix Figure A.1). The silhouette score peaks at $K = 4$ (0.357). The Calinski–Harabasz index favours $K = 3$, but that solution merges Dismissal and Grief into one negative cluster, collapsing the distinction that RQ2 examines. $K = 5$ splits the Ambivalent cluster into two subgroups without an interpretable difference. We therefore use $K = 4$. A silhouette of 0.357 indicates moderate overlap, as expected for continuous emotional responses, so we interpret clusters as ideal-typical profiles rather than sharply separated groups.

**Behaviour at low counts.** Because of zero replacement, a post with few reactions lies close to the CLR origin. Assigning synthetic posts to the exact cluster centroids (Appendix Table A.1) shows that every single-reaction post falls in Ambivalent, whereas two Haha or two Sad reactions suffice for Dismissal or Grief and Love-only posts need four for Solidarity. Sparse posts therefore carry little information, which motivates the minimum-count check below.

**Robustness checks.** We repeat the clustering (i) requiring at least 10 differentiated reactions per post, (ii) with pseudocounts of 0.1, 0.01, and 0.001, (iii) in ILR coordinates (orthonormal Helmert basis), (iv) with a Gaussian mixture model (GMM) in CLR space, (v) on bootstrap resamples, and (vi) with Like included in the composition. We compare partitions with the adjusted Rand index (ARI) and, separately, test whether the central temporal contrast (Dismissal highest in 2022, Grief highest in 2024) is preserved.

## 4.3 Distributional Analysis

We cross-tabulate cluster membership against page country (top 10), page category (top 10), post type, and crisis period, reporting $\chi^2$ tests, Cramér's $V$, and Pearson residuals $r = (O - E)/\sqrt{E}$. Because posts are nested in pages and $n$ is large, we treat $V$ and residual magnitudes as the evidence and $p$-values as descriptive; with four clusters, $V$ of 0.06, 0.17, and 0.29 marks small, medium, and large effects (Cohen, 1988). Periods are compared on monthly cluster shares (53 months) with Kruskal–Wallis and Bonferroni-corrected Mann–Whitney tests. To account for posts nested within pages, we also fit a page-clustered multinomial generalized estimating equation (GEE; exchangeable working correlation; reference outcome Solidarity, reference period COVID-19), and page fixed-effects regressions of log(shares + 1) and log(comments + 1) on cluster, controlling for period and post type, with page-clustered standard errors.

### 4.4 Can Text Recover the Profile?

Three studies examine RQ3; Studies A and B differ in model and protocol, and Study C is a matched ablation.

**Study A: reaction-informed LLM and human labels (*n* = 800).** A stratified sample of 200 posts per cluster was labelled by Claude Opus 4.7 using a two-step, content-first protocol: the model assigned a provisional label from the post text, page name and category, and page country, and then checked it against the reaction percentages, revising only on a clear discrepancy. The prompt listed six countries classified as fragile or conflict-affected (Iraq, Sudan, Syria, Lebanon, Libya, Yemen (Hill et al., 2025)) as low-trust contexts. A paid native Arabic-speaking annotator labelled the same 800 posts independently, without seeing the LLM labels, using the same content-first guidelines with access to the text, metadata, and reactions (Appendix A). The LLM–human comparison is therefore an independent model–human evaluation. The comparison between human labels and *k*-means profiles is not fully independent: the annotator saw reactions, and the guidelines include reaction thresholds derived from the cluster profiles. It measures whether the profiles are *interpretable* to a native reader rather than independent accuracy.

**Study B: reaction-blind prediction (*n* = 6,225).** A balanced sample of 10,000 posts (2,500 per cluster) was labelled zero-shot by Fanar (FANAR-C-2-27B; Fanar Team, 2025), an Arabic-centric LLM, from the post text, page name and category, and page country only, with no reaction information. The prompt (Appendix B) defines each label by the audience response the content is likely to draw and includes the same low-trust country list. The model declined 3,775 posts (37.8%) under its content-safety filters, leaving 6,225 labelled posts. We compare these labels with the *k*-means assignments; per-cluster agreement is therefore recall with respect to the reaction-based profile.

**Study C: matched ablation (*n* = 294).** To isolate reaction access, Fanar labelled a further 328 posts twice under otherwise identical prompts: with text and metadata only, and with the six reaction percentages added (deterministic decoding, temperature 0, top-*p* 1; Appendix B). The 294 posts labelled in both conditions form the paired set; we compare conditions against the *k*-means labels with an exact McNemar test and, because 34 posts were refused in the reaction-aware condition, report inverse-probability-weighted accuracies and bounds.

## 5 Results

### 5.1 Four Reaction Profiles (RQ1)

Love is the most frequent differentiated reaction (mean share 35.1%), but Haha and Sad + Angry together account for 43.0% of the mean composition, a heterogeneity that aggregate sentiment would hide. Table 1 shows the four clusters. Three are anchored by a single reaction: **Solidarity** by Love (67.7%), **Dismissal** by Haha (58.0%), and **Grief** by Sad (60.1%); the **Ambivalent** cluster, half of the corpus, has no reaction above 28%. The names are shorthand for reaction-share profiles, not measures of individual emotion: a Haha-dominant profile may reflect ridicule, sarcasm, or amusement, a Love-dominant one endorsement or sympathy, and a flat one a mix of responses across audience members. Table 2 shows that the profiles do not follow tone: the Solidarity and Dismissal examples are equally factual economic news.

| Cluster | Love | Care | Wow | Haha | Sad | Angry | Share |
|---|---|---|---|---|---|---|---|
| Ambivalent | 27.9 | 15.2 | 13.5 | 15.5 | 13.8 | 14.0 | 50.3 |
| Solidarity | **67.7** | 12.0 | 4.4 | 7.0 | 4.5 | 4.5 | 26.6 |
| Dismissal | 15.4 | 6.4 | 6.8 | **58.0** | 6.0 | 7.5 | 15.9 |
| Grief | 9.2 | 6.2 | 7.0 | 7.2 | **60.1** | 10.4 | 7.2 |

**Table 1:** Reaction profiles: mean % of differentiated (non-Like) reactions per post, and each cluster's share of the 74,241 posts (*n* = 37,339 / 19,715 / 11,841 / 5,346).

| Profile | Post (page country) and gloss |
|---|---|
| Solidarity | الصادرات الزراعية تحقق رقماً قياسياً بـ ٦.٥ مليار دولار(EG)<br>*Agricultural exports reach a record $6.5 billion.* |
| Dismissal | قرض جديد لتونس بقيمة ٢٣٣ مليون دولار لتأمين الأمن الغذائي والطاقي(TN)<br>*A new $233 million loan for Tunisia to secure food and energy security.* |
| Grief | الأمم المتحدة تحذر من أسوأ أزمة غذاء يشهدها السودان(GB)<br>*The UN warns of the worst food crisis Sudan has witnessed.* |
| Ambivalent | رئيس الحكومة السورية: روسيا دعمتنا باحتياطيات كافية من القمح(RU)<br>*Syrian Prime Minister: Russia has supported us with sufficient wheat reserves.* |

**Table 2:** One post per profile. More examples are given in Appendix Table A.6.

## 5.2 Distribution by Source and Context (RQ2)

**Page category.** Category is the strongest correlate of profile ($\chi^2$ = 6,306.5, df = 27, $V$ = 0.215). Government organisations and politicians are over-represented in Solidarity ($r$ = +27.9, +26.0) and under-represented in Dismissal and Grief. News-media companies show the reverse (Solidarity −28.8; Dismissal +16.7; Grief +16.8), and TV channels show the largest Dismissal excess (+19.8). Non-profits (+12.5), personal pages (+13.3), and blogs (+13.9) also lean towards Solidarity (Appendix Figure A.3).

| Page country | Amb. | Dism. | Grief | Solid. |
|---|---|---|---|---|
| Algeria (DZ) | **−11.3** | −3.8 | **−13.8** | **+24.2** |
| Egypt (EG) | +4.5 | **−26.5** | **−10.1** | **+19.0** |
| Iraq (IQ) | −3.9 | **+24.8** | **−19.4** | −4.9 |
| Jordan (JO) | +2.2 | −2.7 | −1.6 | −0.2 |
| Lebanon (LB) | +4.5 | −3.6 | **+13.9** | −9.7 |
| Mauritania (MR) | +6.1 | **+13.6** | −6.8 | **−15.3** |
| Syria (SY) | +7.8 | −4.2 | **+10.0** | **−11.8** |
| Tunisia (TN) | −6.3 | +2.6 | +2.4 | +5.2 |
| Turkey (TR) | −1.1 | −3.2 | **+30.8** | **−10.4** |
| UAE (AE) | −3.3 | **+10.6** | **+30.3** | **−17.9** |

**Table 3:** Pearson residuals of cluster membership by page-administration country (top 10 by volume). Bold: $|r| > 10$. Country is where the page is administered, not where its audience is located.

**Page country.** Country is a comparable correlate ($\chi^2$ = 6,795.7, df = 27, $V$ = 0.200; Table 3). Three patterns stand out. (i) Iraq has the strongest Dismissal excess (+24.80) together with a Grief deficit (−19.36). (ii) Egypt and Algeria lean towards Solidarity (+19.01, +24.20), and Egypt is strongly under-represented in Dismissal (−26.51). (iii) UAE- and Turkey-administered pages show the largest Grief excess (+30.32, +30.83). In the UAE this traces to pan-Arab news channels headquartered there (e.g., Al Arabiya, Asharq News), whose coverage spans crises in Sudan, Syria, and Gaza. This illustrates why page country indexes the production environment rather than domestic sentiment.

**Post type.** The association is weaker ($V$ = 0.105). Live video (47.0%) and native video (31.7%) have the highest Solidarity shares. Link posts, which mostly point to news articles, have the highest Grief share (14.3%) and the lowest Solidarity share (7.9%; Appendix Figure A.4).

**Crisis periods.** Profile shares differ across periods ($\chi^2$ = 2,735.8, df = 9, $V$ = 0.111; Table 4, Figure 1). Ambivalent dominates throughout (47.6–52.9%). Solidarity is stable (25.4–28.2%; Kruskal–Wallis on monthly shares $H$ = 4.68, $p$ = .197). Dismissal and Grief vary significantly across periods (both $p < .001$). Dismissal roughly doubles from the COVID-19 period (10.7%) to 2022 (22.3%), with a sharp rise after February 2022. Grief reaches 14.0% in 2024, against 3.9% in 2022; pairwise tests show that 2024 differs from both 2022 and 2023. The monthly series shows the Grief rise beginning in late 2023. The page-clustered GEE (74,241 posts, 15,144 pages) confirms both shifts relative to COVID-19: higher odds of Dismissal in 2022 ($\beta$ = 0.781, OR = 2.18, $p < .001$) and of Grief in 2024 ($\beta$ = 0.293, OR = 1.34, $p < .001$).

| Period | Amb. | Dism. | Grief | Solid. |
|---|---|---|---|---|
| COVID-19 (2020–21) | 52.9 | 10.7 | 9.8 | 26.7 |
| Ukraine/wheat (2022) | 48.3 | **22.3** | 3.9 | 25.4 |
| Post-crisis (2023) | 52.8 | 13.4 | 6.2 | 27.6 |
| Gaza (Jan–Jul 2024) | 47.6 | 10.3 | **14.0** | 28.2 |

**Table 4:** Cluster shares (% of posts) by crisis period.

| | **Study A** (reaction-informed) | | **Study B** |
|---|---|---|---|
| Profile | LLM–H | KM–H | Blind–KM |
| Solidarity | 100.0 | 100.0 | 57.4 |
| Grief | 89.5 | 89.0 | 83.8 |
| Dismissal | 73.4 | 88.9 | **3.4** |
| Ambivalent | 55.4 | 73.1 | 47.6 |
| Overall | 79.8 | 87.9 | 43.4 |
| 95% CI | [76.9, 82.4] | [85.5, 90.0] | [42.2, 44.6] |
| Cohen's $\kappa$ | 0.728 | 0.838 | 0.244 |
| 95% CI | [.69, .77] | [.81, .87] | [.23, .26] |
| $n$ | 800 | 800 | 6,225 |

**Table 5:** Agreement (%) by profile. KM: $k$-means profile; H: independent native-speaker label; LLM: Claude Opus 4.7 with reactions (Study A); Blind: Fanar with text and source metadata, no reactions (Study B). Study B per-profile values are recall against KM. Intervals: Wilson for agreement; large-sample normal approximation for $\kappa$. The two studies differ in model, protocol, and reference, so the columns are not a controlled ablation.

## 5.3 Robustness of the Profiles

The temporal contrast holds under every check (Appendix Table A.2). Requiring 10 differentiated reactions preserves the 2022 Dismissal and 2024 Grief peaks. ILR coordinates give identical assignments (ARI = 1.000), as expected from the isometry. Bootstrap refits are highly stable (median ARI = 0.989), and including Like changes little (ARI = 0.909; period shares within 2.2 points). Post-level boundaries are less stable: smaller pseudocounts reassign many posts (ARI 0.22–0.29) and a GMM gives a coarser partition (ARI = 0.359), yet in all of them Dismissal is highest in 2022 and Grief in 2024. We therefore treat the profiles as ideal-typical audience-response regimes whose aggregate dynamics are robust, rather than as uniquely determined classes.

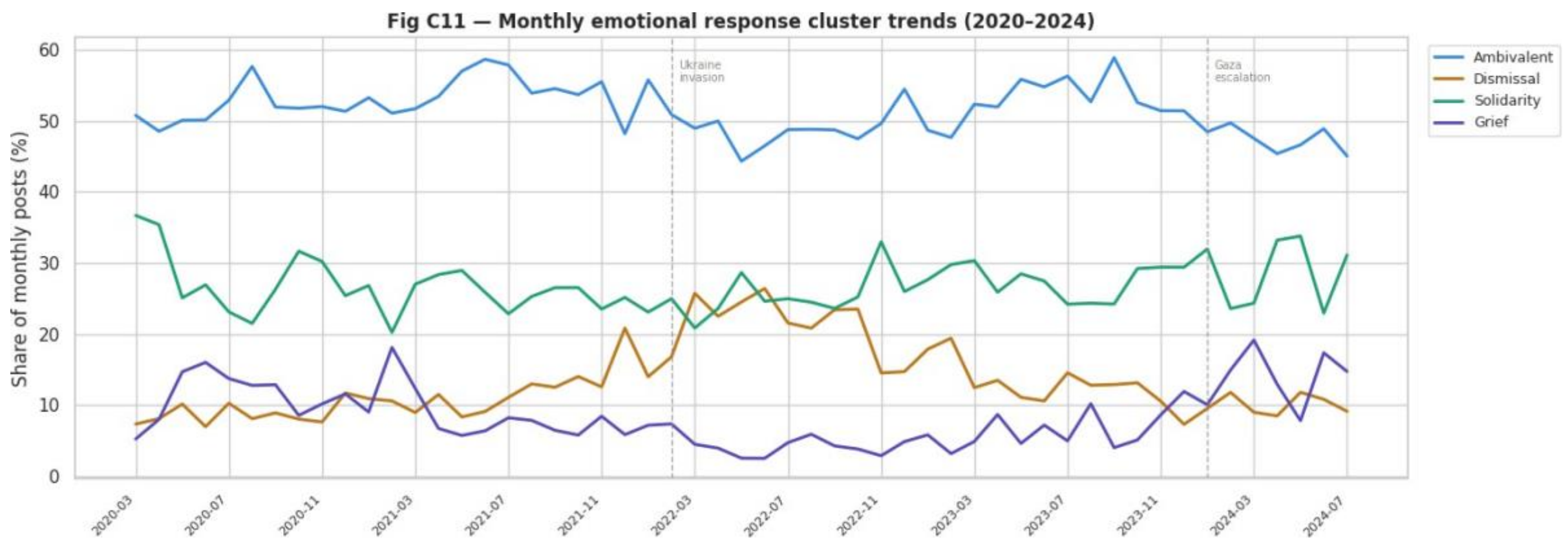

**Figure 1:** Monthly cluster shares, March 2020 to July 2024. Dashed lines mark the invasion of Ukraine (February 2022) and the Gaza escalation (January 2024). Dismissal rises in 2022; Grief rises from late 2023 into 2024.

## 5.4 Engagement by Profile

Engagement differs across clusters (Kruskal–Wallis, all $p < .001$). Solidarity posts have the most shares and comments (medians 6 and 19) and are the only ones whose median CrowdTangle Overperforming Score is positive (+1.14, i.e., 14% above the page's own benchmark for posts of that type);[3] the others underperform (Ambivalent −2.02, Dismissal −1.38, Grief −1.70). Dismissal posts draw many comments (median 17) but few shares (median 2). Page fixed-effects models confirm this within pages: relative to Ambivalent posts from the same page, Solidarity has the largest association with shares ($\beta = 0.728$) and Dismissal with comments ($\beta = 1.101$; both $p < .001$). Affiliative posts travel; dismissive posts generate discussion. These are page-relative associations, not measures of exposure or algorithmic ranking.

## 5.5 Reaction-Blind Prediction of the Profile (RQ3)

Table 5 reports both studies. In Study A, the independent human labels agree with the reaction-informed LLM on 79.8% of posts ($\kappa = 0.728$) and with the $k$-means profile on 87.9% ($\kappa = 0.838$). All three agree on every Solidarity post. The LLM is weakest on Ambivalent (55.4%) and Dismissal (73.4%), the two profiles in which audience response depends most on context. Agreement between humans and *k*-means is lowest for Ambivalent (73.1%), whose flat profiles a reader can attribute to several kinds of content. Because the annotator saw reactions, the human–*k*-means figures show that the profiles are legible to a native reader; they are an upper bound rather than an independent accuracy estimate.

[3] A score of +2 means twice the expected interactions and −2 half; see the CrowdTangle help article "How do you calculate overperforming scores?".

Study B withholds reactions. Among the 6,225 labelled posts, agreement with the reaction-based profile is 43.4% ($\kappa = 0.244$); counting the 3,775 refusals as errors, agreement over all 10,000 attempted posts is 27.0%. Among labelled posts, the mean per-profile recall is 48.1% against a four-way chance level of 25%. Recall varies sharply by profile (macro-F1 0.43; Appendix Table A.4). Grief is largely recoverable (recall 83.8%, precision 79.1%): posts that report suffering are typically met with Sad. Solidarity (57.4%) and Ambivalent (47.6%) are partially recoverable. Dismissal is almost never recovered (3.4%), even though the prompt names as low-trust several countries in which Dismissal concentrates. The confusion matrix shows why. Fanar assigns Dismissal to only 98 of 6,225 posts (1.6%), although 29.4% of the labelled posts carry that profile; when it does, it is usually right (precision 64.3%). It labels the remaining Dismissal posts Ambivalent (52.9%) or Solidarity (39.6%). Dismissal posts are typically official announcements, budgets, and committee meetings whose wording is neutral or positive (Appendix Table A.5). The model reserves Dismissal for a few posts with overt cues and otherwise reads such posts at face value. Its correct Dismissal predictions show what those cues are: the derision is written into the post itself, as when a lawmaker calls a provision of Iraq's Food Security Law a "heresy" (بدعة), a commentator calls the law's wording "clownish" (بهلوانية), or a post ends in laughing emoji. Of the 63 correct predictions, 34 come from Iraqi pages. Posts whose derision lies only in the audience's response, such as neutral announcements of budgets and meetings, are labelled Ambivalent or Solidarity instead.

Showing Fanar the reactions does not fix this (Study C; Appendix Table A.3). On the 294 paired posts, accuracy is 51.4% without reactions and 49.0% with them (macro-F1 0.480 vs. 0.465; exact McNemar $p = .189$). The model changes only 36 predictions (12.2%), 7 for the better and 14 for the worse. Dismissal recall is 0.0% without reactions and 3.8% with them, against 92.9% for Grief in both conditions. Weighting for the uneven refusals (19.2% of Dismissal posts against 3.4–7.2% for the others) preserves the direction (49.0% vs. 46.9%), although bounds for the refused cases span 43.9–54.3%. Even a reaction vector with 60% Haha does not lead the model to choose Dismissal.

Two checks bear on this result. *Refusals*: had every refused post been labelled, overall agreement would lie between 27.0% (all wrong) and 64.8% (all correct). Refusals were also uneven: 58.6% of sampled Grief posts were refused, against 26.7–37.2% for the other profiles, so Grief recall over all 2,500 sampled posts is at least 34.7%. Even if every refused Dismissal post had been labelled correctly, Dismissal recall would be at most 29.2%. *Low counts*: some Dismissal posts owe their label to two Haha reactions (§4.2), but the independent annotator assigned Dismissal to 88.9% of sampled Dismissal posts, close to the rate for Grief (89.0%).

### 5.6 Supervised Baselines

Is Dismissal hard to predict from text, or is the zero-shot model simply miscalibrated? We trained logistic-regression classifiers (balanced class weights) on the 6,225 Study B posts, using the same input text (the first 400 characters of each post) and the reaction-based profiles as labels. Features are word unigrams and bigrams plus character 2–5-grams (TF-IDF), one-hot source metadata (page category, page country, post type), or both. We evaluate with five-fold cross-validation grouped by page, so no page appears in both training and test folds, and score every post once against the same reference as Fanar.

Table 6 and Figure 2 show that a text-only model recovers Dismissal for 60.0% of posts (95% bootstrap CI 57.8–62.3), against 3.4% for Fanar, and improves macro-F1 from 43.1 to 55.8 without losing much on Grief. The gain holds outside Iraq (Dismissal recall 54.5%, against 75.4% for Iraqi posts) and under a temporal split in which models are trained on 2020–2022 and tested on 2023–2024 posts from pages unseen in training (macro-F1 53.6; Dismissal recall 48.1%). Metadata alone already recovers half of the Dismissal posts, confirming that source matters, but text adds most of the gain. The most predictive Dismissal features name officials and institutions (وزير 'minister', قانون 'law', الحكومة 'government', and named heads of government), whereas Grief features describe suffering (غزة 'Gaza', انعدام 'lack', نقص الغذاء 'food shortage'). Dismissal is thus predictable from *who* a post is about rather than from its tone.

| | | | Dismissal | | Grief |
|---|---|---|---|---|---|
| Model | M-F1 | κ | R | P | R |
| Fanar, zero-shot | 43.1 | .244 | 3.4 | 64.3 | 83.8 |
| LR, metadata only | 41.0 | .236 | 51.1 | 47.2 | 53.6 |
| LR, text only | 55.8 | .385 | 60.0 | 57.6 | 78.1 |
| LR, text + metadata | **57.3** | **.409** | **63.5** | 60.5 | 75.7 |

**Table 6:** Reaction-blind prediction on the 6,225 Study B posts (%). LR models use five-fold page-grouped cross-validation. M-F1: macro-F1; R/P: recall/precision. Majority-class accuracy is 29.4%.

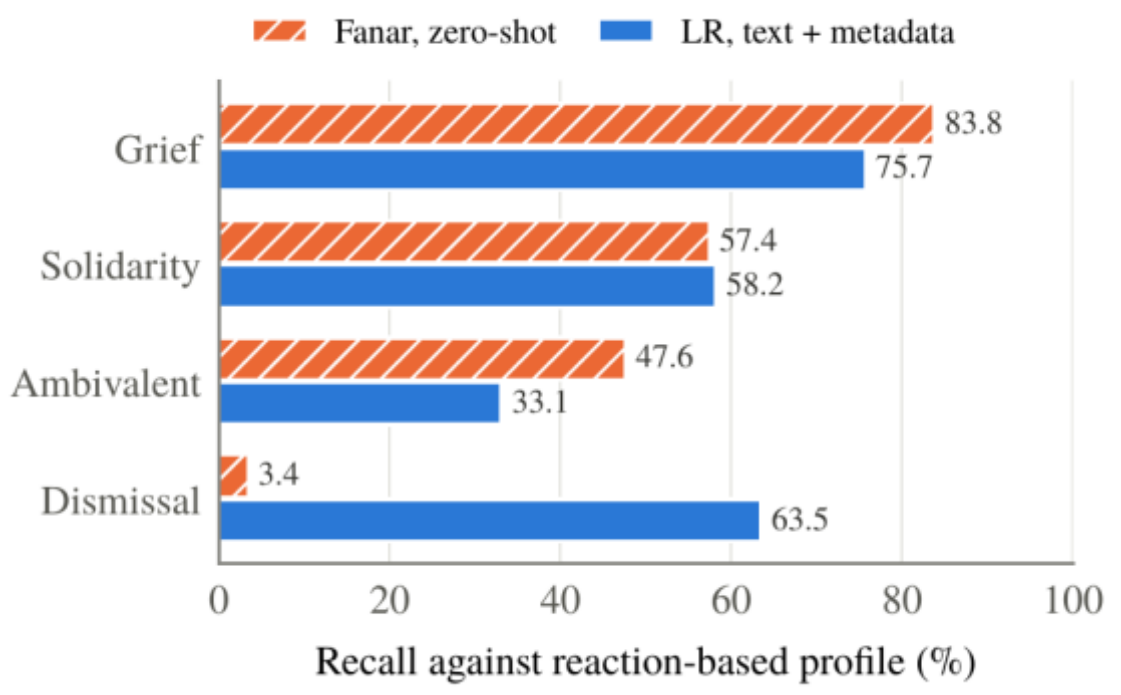

**Figure 2:** Per-profile recall of the zero-shot LLM and the page-disjoint supervised baseline on the same 6,225 posts. Training on reaction-derived labels recovers Dismissal, which the zero-shot model almost never predicts.

## 6 Discussion

**What reaction-derived labels teach.** The contrast between Study B and the supervised baselines is the central NLP result. Distress is stated in the post, and both the zero-shot LLM and the trained models recover Grief. Derision is not stated; it attaches to particular actors and institutions. A model trained on reaction-derived labels learns that association from text and recovers most Dismissal posts, whereas the zero-shot LLM, which reads tone at face value, almost never assigns the label (1.6% of posts), even though its rare Dismissal predictions are usually correct, and does not assign it even when shown the reactions (Study C). Since a simple lexical model recovers what the LLM misses, the gap reflects calibration rather than missing signal. LLM labelling pipelines, now widely used to annotate Arabic social media, would therefore under-label this audience response unless they are adapted with reaction-derived supervision of the kind FoodReact-AR provides. Fine-tuned Arabic encoders and few-shot prompting are natural next steps under the protocol below.

**An evaluation protocol for audience-profile prediction.** The released labels define a four-way task: predict a post's reaction profile from its text and source metadata. To keep results comparable and free of the confounds above, we recommend (1) restricting evaluation to posts with at least 10 differentiated reactions, where profiles are not decided by single reactions; (2) page-disjoint train, development, and test splits, so that models cannot memorise page-specific audience habits; (3) an additional temporal split (train on 2020–2022, test on 2023–2024) for prospective use; (4) macro-F1 as the primary metric, with Dismissal recall reported separately as the hardest case; and (5) counting refusals or unparseable outputs as errors and reporting their rate. The baselines in Table 6 give reference points for this task.

**Crisis periods: association, not mechanism.** The 2022 Dismissal peak and the 2024 Grief peak are consistent with framing accounts in which economic and geopolitical frames invite blame and scepticism while humanitarian frames invite empathy (Oleskog Tryggvason and Shehata, 2024). They survive page-level dependence and the robustness checks, but we did not measure framing, and source composition changes over time. Iraq's parliament passed the Emergency Law for Food Security and Development on 8 June 2022,[4] and our Dismissal examples include Iraqi posts about this law (Appendix Tables A.6 and A.5), so part of the 2022 peak may reflect Iraqi domestic politics rather than the wheat shock. Likewise, the 2024 Grief peak coincides with Gaza coverage by pan-Arab news pages, which are strongly over-represented in Grief (Table 3). We therefore present the contrast as a robust period-level association, not as evidence of a framing mechanism.

[4]https://www.al-monitor.com/originals/2022/06/iraqi-parliament-passes-urgent-food-security-bill

**Solidarity.** Solidarity is the most stable profile across periods and the only one with a positive median Overperforming Score: the median Love-dominant post exceeds its own page's interaction benchmark, which bears on whether non-negative emotion can travel as well as anger (Gerbaudo et al., 2023). Its within-page sharing advantage shows that this is not only a page effect, although government and political pages dominate the profile, and reaction-blind recall for Solidarity is only 57.4%, suggesting that it too partly reflects the audience's relation to the source.

## 7 Conclusion

Treating Facebook reactions as compositions yields four interpretable audience profiles in Arabic food-insecurity discourse. These profiles vary with source type, page country, and crisis period, and the 2022 Dismissal and 2024 Grief peaks survive page-level models and robustness checks. The clearest case is the Haha-dominant Dismissal profile, which concentrates on official food-security messaging: a zero-shot Arabic LLM almost never predicts it, even when shown the reactions, while a simple classifier trained on reaction-derived labels recovers most of it from mentions of officials and institutions. For Arabic NLP, audience response should be treated as a separate prediction target learned from reactions, not inferred from text sentiment; FoodReact-AR supports that work.

## Data Release

FoodReact-AR has four layers: (1) the 74,241 analytical posts with text, page metadata, reaction counts and shares, CLR coordinates, and cluster labels; (2) the 10,000-post Fanar sample with labels for 6,225 posts and refusal flags; (3) the 800-post sample with reaction-informed LLM and independent native-speaker labels; and (4) the 328-post ablation with paired labels and refusals. Folds, splits, prompts, and baseline code accompany it. To respect platform terms, access is granted on request for non-commercial research under a data use agreement that prohibits re-identification and redistribution.

## Limitations

**Clustering robustness.** The temporal contrast is robust (§5.3), but post-level assignments depend on the zero-replacement constant (ARI 0.22–0.29 for smaller pseudocounts), and GMM agreement with *k*-means is moderate. The primary analysis keeps posts with a single differentiated reaction to maximise coverage; such posts always fall in Ambivalent, so that profile should be read cautiously. Profiles are ideal-typical regimes rather than natural classes.

**Annotation design.** Study A uses a single human annotator. The annotator labelled without seeing the LLM labels, but saw reactions and used guidelines whose thresholds were derived from the cluster profiles, so the human–*k*-means agreement is partly circular. A single annotator also does not establish inter-annotator reliability. Study B differs from Study A in model, prompt, and reference labels, so the drop in agreement cannot be attributed to the removal of reaction access alone. Fanar declined 37.8% of posts in Study B, concentrated in Grief (58.6%), and refused Dismissal posts disproportionately in Study C; weighting preserves the Study C result, but its bounds remain wide (43.9–54.3%). Both describe the posts the model would label, and both use a single generative model.

**Country-level trust.** Both prompts and the guidelines name six countries as low-trust contexts, following fragile and conflict-affected classifications. That classification is not a validated measure of trust in a particular institution, page, or message, and it may have steered annotators and models towards Dismissal for those countries. The country associations in Table 3 are observed; their interpretation as institutional distrust is a hypothesis. We keep the prompts unchanged in Appendix B for reproducibility.

**Baselines.** The supervised baselines use the 6,225 posts Fanar labelled, a sample shaped by its refusals, and post text truncated to 400 characters. They are lexical models, which may partly exploit outlet signatures and topical cues rather than modelling derision itself; page-grouped and temporal splits limit but do not remove this. Neural Arabic encoders were not evaluated.

**Statistical inference.** The page-clustered GEE and page fixed-effects models address repeated posts within pages, but they are not multilevel models of pages within countries, and the period estimates are not adjusted for shifts in country or category composition.

**Measurement scope.** Page country is not audience location. Reaction shares describe aggregate audience response, not individual emotional states. The Overperforming Score measures page-relative interactions, not exposure or algorithmic ranking. Retrieval used three Modern Standard Arabic phrases, so dialectal discussions of food insecurity are under-represented and the recall of the collection is unknown. The Care reaction was rolled out in spring 2020, so for the earliest posts some zero Care counts may reflect its unavailability.

## Ethical Considerations

**Data collection and privacy.** All data were collected from public Facebook pages through CrowdTangle, Meta's research tool, in accordance with its terms of use. The pages belong to organisations, media outlets, and public figures. We collected no private content, user comments, user profiles, or individual-level reaction data; reactions are used only as aggregate counts per post. The data contain no personally identifiable information about private individuals. The study protocol was reviewed and approved by the authors' institutional review board (IRB).

**Annotation.** The human annotator was a paid research assistant compensated at local rates and informed of the study's purpose.

**Potential misuse.** Food insecurity, conflict, and institutional distrust are sensitive topics. The profiles describe aggregate audience reactions to public content; they should not be used to judge the credibility of specific institutions, to profile communities, or to optimise manipulative messaging.

## References

Ibrahim Abu Farha and Walid Magdy. 2020. From Arabic sentiment analysis to sarcasm detection: The ArSarcasm dataset. In *Proceedings of the 4th Workshop on Open-Source Arabic Corpora and Processing Tools, with a Shared Task on Offensive Language Detection*, pages 32–39, Marseille, France. European Language Resources Association.

John Aitchison. 1986. *The Statistical Analysis of Compositional Data*. Chapman and Hall, London.

Sawood Anwar and Fabio Giglietto. 2024. Facebook reactions in the context of politics and social issues: a systematic literature review. *Frontiers in Sociology*, 9:1379265.

Tadeusz Caliński and Jerzy Harabasz. 1974. A dendrite method for cluster analysis. *Communications in Statistics*, 3(1):1–27.

Jacob Cohen. 1988. *Statistical Power Analysis for the Behavioral Sciences*, 2 edition. Lawrence Erlbaum Associates, Hillsdale, NJ.

Yassine El Moudene, Jaafar Idrais, Rida El Abassi, and Abderrahim Sabour. 2023. Gender-based analysis of user reactions to facebook posts. *Big Data Mining and Analytics*, 7(1):75–86.

Fanar Team. 2025. Fanar: An Arabic-centric multimodal generative AI platform. *Preprint*, arXiv:2501.13944.

Cole Freeman, Hamed Alhoori, and Murtuza Shahzad. 2020. Measuring the diversity of Facebook reactions to research. *Proceedings of the ACM on Human-Computer Interaction*, 4(GROUP):1–17.

Cole Freeman, Mrinal Kanti Roy, Michele Fattoruso, and Hamed Alhoori. 2019. Shared feelings: Understanding Facebook reactions to scholarly articles. In *Proceedings of the ACM/IEEE Joint Conference on Digital Libraries (JCDL)*.

Paolo Gerbaudo, Ciro Clemente De Falco, Giulia Giorgi, Silvia Keeling, Antonia Murolo, and Federica Nunziata. 2023. Angry posts mobilize: Emotional communication and online mobilization in the facebook pages of western european right-wing populist leaders. *Social Media+ Society*, 9(1):20563051231163327.

Stephan J Goetz, Connor Heaton, Muhammad Imran, Yuxuan Pan, Zheng Tian, Claudia Schmidt, Umair Qazi, Ferda Ofli, and Prasenjit Mitra. 2023. Food insufficiency and twitter emotions during a pandemic. *Applied Economic Perspectives and Policy*, 45(2):1189–1210.

Michael Greenacre, Eric Grunsky, John Bacon-Shone, Ionas Erb, and Thomas Quinn. 2023. Aitchison's compositional data analysis 40 years on: A reappraisal. *Statistical Science*, 38(3):386–410.

Abdo Hassoun, Khawla Al-Muhannadi, Hussein F Hassan, Ahmed Hamad, Khaoula Khwaldia, Mohamed Buheji, and Ayoub Al Jawaldeh. 2024. From acute food insecurity to famine: how the 2023/2024 war on gaza has dramatically set back sustainable development goal 2 to end hunger. *Frontiers in Sustainable Food Systems*, 8:1402150.

Samuel Hill, Jeetendra Khadan, and Peter Selcuk. 2025. Fragile and conflict-affected situations. *Prospects*.

Yi-Juan Hu and Glen A Satten. 2023. Compositional analysis of microbiome data using the linear decomposition model (ldm). *Bioinformatics*, 39(11):btad668.

Sahar Khamis and Felicity Sena Dogbatse. 2024. The gaza war coverage: The role of social media vs. mainstream media. *IEMed Mediterranean yearbook*, pages 295–300.

Chantal Le Mouël, Agneta Forslund, Pauline Marty, Stéphane Manceron, Elodie Marajo-Petitzon, Marc-Antoine Caillaud, Patrice Dumas, and Bertrand Schmitt. 2023. Can the middle east-north africa region mitigate the rise of its food import dependency under climate change? *Regional Environmental Change*, 23(2):52.

Josep A. Martín-Fernández, Carles Barceló-Vidal, and Vera Pawlowsky-Glahn. 2003. Dealing with zeros and missing values in compositional data sets using nonparametric imputation. *Mathematical Geology*, 35(3):253–278.

Gina M Masullo. 2022. Facebook reactions as heuristics: Exploring relationships between reactions and commenting frequency on news about covid-19. *First Monday*.

Marieke Meeske, Frans Cruijssen, Chris van der Lee, and Emiel Krahmer. 2025. The role of language technology and artificial intelligence in food security policymaking. *Discover Sustainability*.

Annika Molenaar, Dickson Lukose, Linda Brennan, Eva L Jenkins, and Tracy A McCaffrey. 2024. Using natural language processing to explore social media opinions on food security: sentiment analysis and topic modeling study. *Journal of Medical Internet Research*, 26:e47826.

Per Oleskog Tryggvason and Adam Shehata. 2024. Success or failure? news framing of the cop26 glasgow summit and its effects on citizens' beliefs about climate change. *The International Journal of Press/Politics*, 29(3):689–709.

John C Paolillo. 2023. The awkward semantics of facebook reactions. *First Monday*.

Chris Pool and Malvina Nissim. 2016. Distant supervision for emotion detection using Facebook reactions. In *Proceedings of the Workshop on Computational Modeling of People's Opinions, Personality, and Emotions in Social Media (PEOPLES)*, pages 30–39, Osaka, Japan. The COLING 2016 Organizing Committee.

Jaber Rahimi, Andrew Smerald, Hassane Moutahir, Mostafa Khorsandi, and Klaus Butterbach-Bahl. 2023. The potential consequences of grain-trade disruption on food security in the middle east and north africa region. *Frontiers in Nutrition*, 10:1239548.

Peter J. Rousseeuw. 1987. Silhouettes: A graphical aid to the interpretation and validation of cluster analysis. *Journal of Computational and Applied Mathematics*, 20:53–65.

Rubaiya Zannat. 2022. Crisis communication in the ukraine-russia conflict: Use of facebook by un humanitarian organizations.

## A Annotation Details

**Annotator.** Human labels were produced by a paid student research assistant from Tunisia, a native Arabic speaker familiar with Modern Standard Arabic and the dialectal variation common in MENA social media. The annotator labelled all 800 posts independently, without access to the LLM labels, was recruited through an academic network, compensated at local student research-assistant rates, and worked on a flexible schedule; the workload was limited to 800 posts.

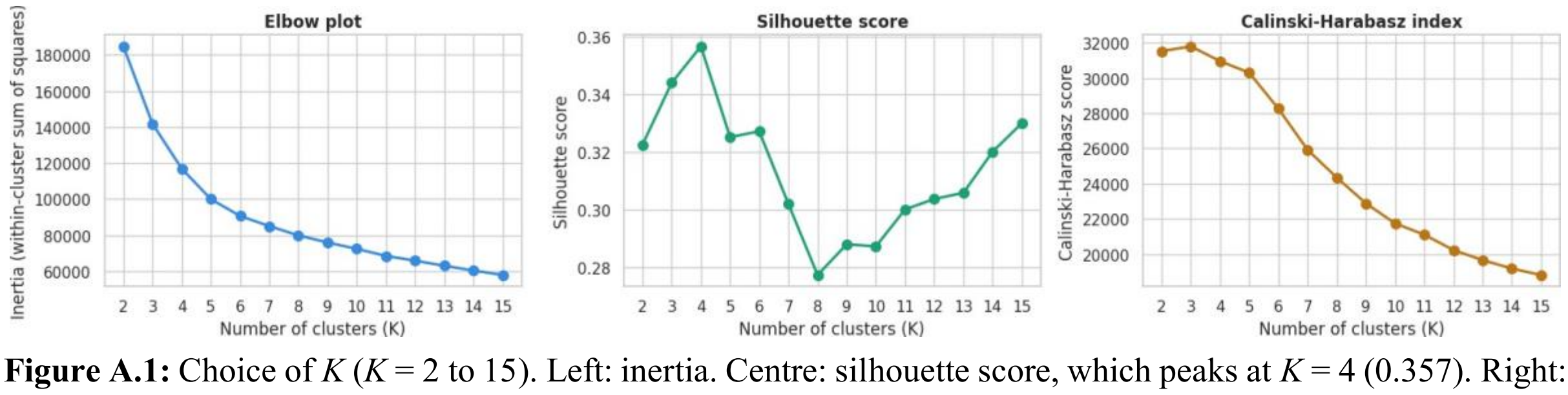

**Figure A.1:** Choice of $K$ ($K = 2$ to 15). Left: inertia. Centre: silhouette score, which peaks at $K = 4$ (0.357). Right: Calinski–Harabasz index.

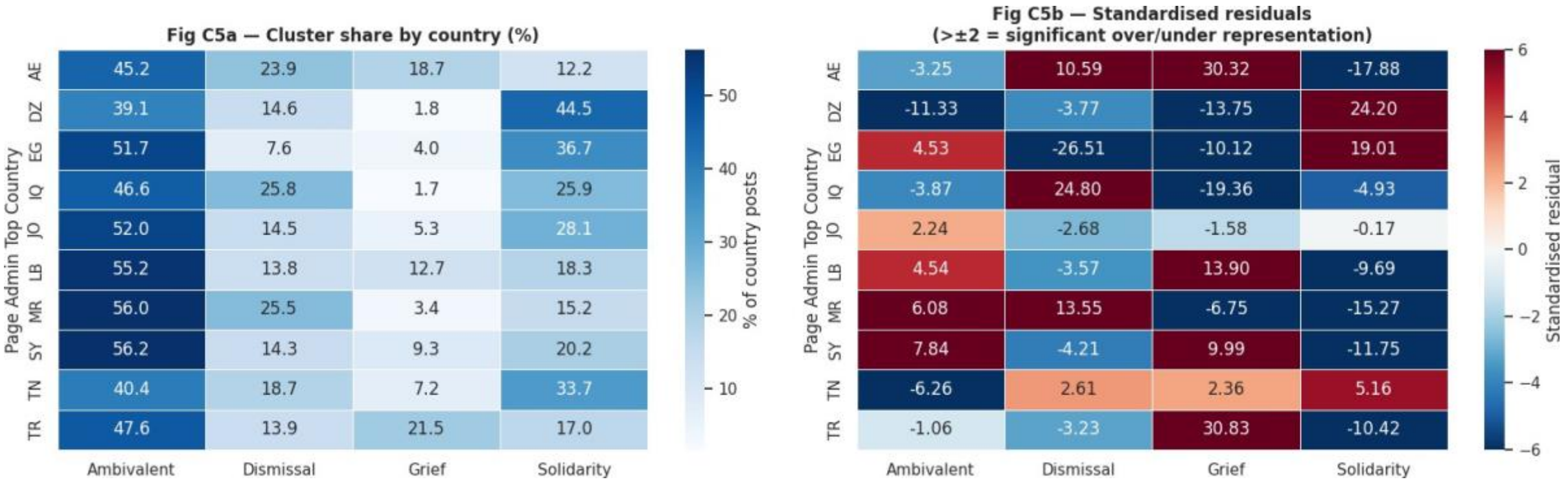

**Figure A.2:** Cluster distribution by page-administration country. Left: % of posts per cluster. Right: Pearson residuals (red: over-representation; blue: under-representation; $|r| > 2$ notable).

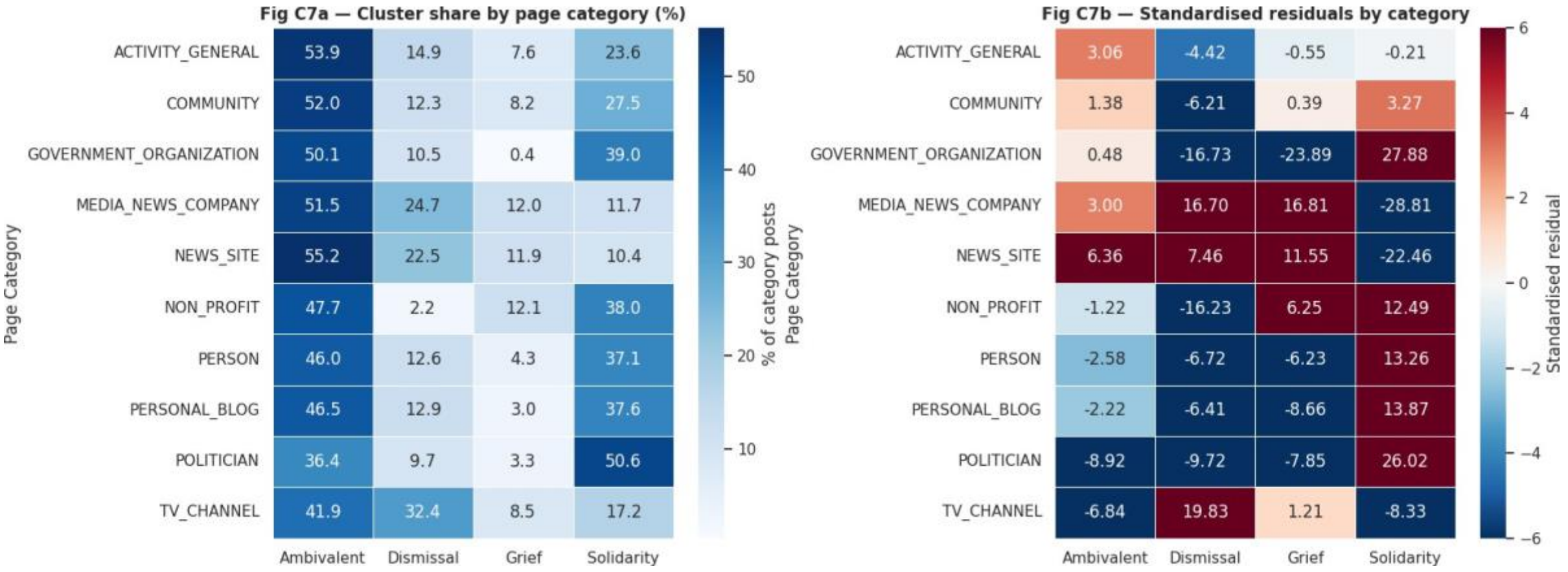

**Figure A.3:** Cluster distribution by page category. Left: % of posts per cluster. Right: Pearson residuals.

| Cluster | Love | Care | Wow | Haha | Sad | Angry |
|---|---|---|---|---|---|---|
| Solidarity | 2.24 | 0.30 | −0.74 | −0.33 | −0.73 | −0.74 |
| Dismissal | 0.26 | −0.56 | −0.49 | 1.89 | −0.67 | −0.44 |
| Grief | −0.20 | −0.58 | −0.53 | −0.52 | 1.96 | −0.13 |
| Ambivalent | 0.51 | −0.06 | −0.15 | −0.03 | −0.13 | −0.13 |

**Table A.1:** CLR-space centroids (rounded) of the four *k*-means clusters. Back-transformed, they correspond to compositions of 72.9% Love (Solidarity), 64.4% Haha (Dismissal), 67.4% Sad (Grief), and a flat profile with 26.9% Love (Ambivalent).

**Guidelines.** The annotator received written guidelines and examples for all four categories, including borderline cases involving sarcasm, political distrust, humanitarian framing, and mixed signals. The guidelines asked for a semantic reading before inspecting reactions and gave particular attention to (1) source credibility in the national political context, (2) ironic or sarcastic uses of Haha, (3) mixed humanitarian and political framing, and (4) reaction distributions without a dominant signal. The following thresholds, derived from the cluster profiles, served as soft heuristics: Haha > 40% suggests Dismissal; Sad > 40% suggests Grief; Love > 50% suggests Solidarity; no reaction above 35% with Love the largest suggests Ambivalent.

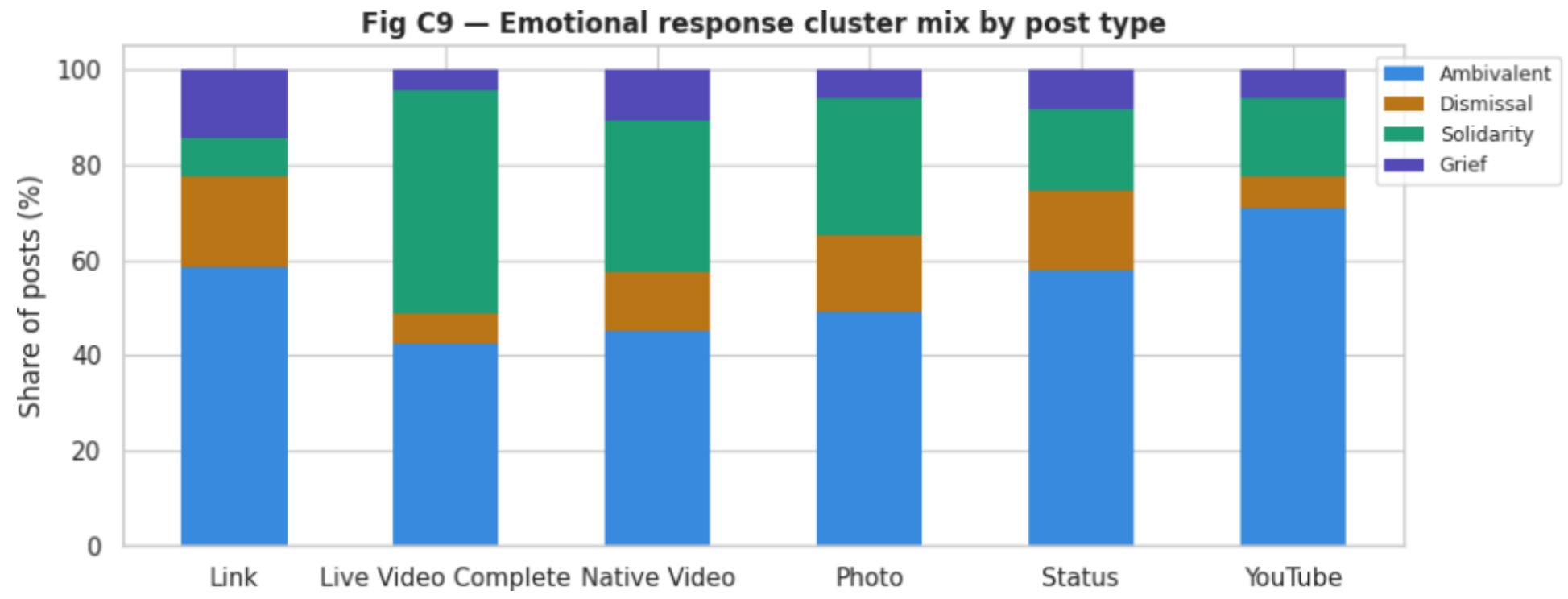

**Figure A.4:** Cluster distribution by post type.

| Specification | ARI | Agree. | Contrast |
|---|---|---|---|
| ≥10 differentiated reactions | n/a | n/a | ✓ |
| ILR coordinates (Helmert) | 1.000 | 100.0 | ✓ |
| Pseudocount 0.1 | 0.290 | 66.2 | ✓ |
| Pseudocount 0.01 | 0.232 | 63.2 | ✓ |
| Pseudocount 0.001 | 0.221 | 62.4 | ✓ |
| Gaussian mixture model | 0.359 | 52.3 | ✓ |
| Bootstrap (median) | 0.989 | 99.6 | n/a |
| Like included | 0.909 | n/a | ✓ |

**Table A.2:** Robustness of the $K = 4$ solution against the primary specification (pseudocount 0.5, CLR, *k*-means). Agree.: % of posts with the same assignment. Contrast: Dismissal highest in 2022 and Grief highest in 2024. Under pseudocounts 0.1/0.01/0.001, the 2022 Dismissal share is 32.6/34.5/34.0% and the 2024 Grief share 24.6/21.6/21.0%; the bootstrap 95% interval for ARI is about 0.967–0.996; with Like included, period shares move by at most 2.17 points.

| | | | | **Recall** | |
|---|---|---|---|---|---|
| Condition | Acc. | M-F1 | $\kappa$ | Dism. | Grief |
| Text + metadata | 51.4 | .480 | .349 | 0.0 | 92.9 |
| + reactions | 49.0 | .465 | .316 | 3.8 | 92.9 |

**Table A.3:** Study C: matched Fanar ablation on the 294 posts labelled in both conditions, against the *k*-means labels (%). Exact McNemar $p = .189$. Solidarity recall rises from 66.7 to 71.6 and Ambivalent recall falls from 58.4 to 40.3.

| | **Fanar prediction (reaction-blind)** | | | | |
|---|---|---|---|---|---|
| Profile | Amb. | Dism. | Grief | Solid. | *n* |
| Ambivalent | **747** | 22 | 116 | 684 | 1,569 |
| Dismissal | 969 | **63** | 74 | 726 | 1,832 |
| Grief | 114 | 3 | **868** | 51 | 1,036 |
| Solidarity | 712 | 10 | 40 | **1,026** | 1,788 |
| Predicted | 2,542 | 98 | 1,098 | 2,487 | 6,225 |
| Precision | 29.4 | 64.3 | 79.1 | 41.3 | |
| Recall | 47.6 | 3.4 | 83.8 | 57.4 | |
| F1 | 36.3 | 6.5 | 81.3 | 48.0 | |
| Refused (%) | 37.2 | 26.7 | 58.6 | 28.5 | |

**Table A.4:** Study B confusion matrix: reaction-based profile (rows) against Fanar's reaction-blind label (columns), with per-profile metrics (%) and the share of each profile's 2,500 sampled posts that the model refused. Macro-F1 = 43.1.

**Observed difficulties.** The most frequent difficulties were sarcastic political commentary and posts that combined humanitarian language with political blame. In several Iraqi, Tunisian, and Sudanese examples, high Haha reflected scepticism towards political actors rather than humour. Dialect-specific phrasing and implicit political references also caused ambiguity.

# B Prompts

## Study A (reaction-informed, two-step).

```
SYSTEM: You are annotating Arabic-language Facebook
posts about food insecurity in the MENA region. Assign
a label based primarily on the SEMANTIC CONTENT of the
post and the SOURCE CONTEXT, then verify whether the
audience reactions are consistent with your assessment.

STEP 1 - Content and context assessment (primary)
Read the post text and consider:
- What is being communicated? (crisis report, official
  announcement, aid delivery, political criticism,
  humanitarian appeal, etc.)
- Who is saying it? (government, NGO, media,
  individual, international institution)
```

| Page / Country | Reaction | Post content (English gloss) | KM | A | H |
|---|---|---|---|---|---|
| **Panel A: full agreement (KM = A = H)** | | | | | |
| Tunisian Agriculture Ministry / TN | Love 74% | Tunisian–Libyan Economic Forum: agreement to revive bilateral economic relations, remove obstacles, and create new investment opportunities. | S | S | S |
| Libya Food Security Committee / LY | Haha 64% | The Supreme Committee for Food Security Strategy holds its second meeting, chaired by the Deputy PM, to review the committee agenda. | D | D | D |
| Al Jazeera Sudan / QA | Sad 75% | AFP, citing the UN: 19 million people in Sudan may face food insecurity in the coming months. | G | G | G |
| Sudanese Agriculture Ministry / SD | Flat | Acting Agriculture Minister inspects flour mills and seed companies to secure wheat and grain supply ahead of the season. | Amb | Amb | Amb |
| **Panel B: LLM diverges (KM = H ≠ A)** | | | | | |
| Iraqi media page / IQ | Haha 40% | Breaking: Diwaniyah food-security-law projects approved by the Planning Ministry, covering eight neighbourhoods including four Sadr districts. | D | Amb | D |
| Iraqi Ministry of Trade / IQ | Flat | Ministry of Trade denies the existence of "carcinogenic Australian wheat", stating that no such term exists in global food standards. | Amb | D | Amb |
| **Panel C: *k*-means diverges (A = H ≠ KM)** | | | | | |
| Iraqi TV channel / IQ | Haha 48% | Finance Committee head defends food-security-law allocations: 12 trillion dinars transferred to provinces; oil revenues insufficient. | D | S | S |
| Libyan community page / LY | Wow 51% | Zliten food-safety inspectors tour a local factory and find unhygienic conditions and products unfit for consumption. | G | Amb | Amb |
| Egyptian Int. Cooperation Ministry / EG | Angry 70% | Minister delivers Egypt's speech at the Korea–Africa Summit, conveying the President's greetings to summit leaders. | Amb | D | D |

**Table A.5:** Study A error analysis. KM: *k*-means; A: reaction-informed LLM (Claude Opus 4.7); H: independent human label. S: Solidarity; D: Dismissal; G: Grief; Amb: Ambivalent. In Panel B the LLM follows surface cues (denial wording read as contestation, policy wording as neutral) and misses source distrust that reactions capture. In Panel C reactions mislead when emotions are mixed (Haha with Love, Wow with Angry) or when the dominant reaction targets the source rather than the topic. Solidarity is absent from Panels B and C because all three agreed on every Solidarity post.

```
- Is the source credible given the country context?
  Governments in Iraq, Sudan, Syria, Lebanon, Libya,
  and Yemen have low institutional credibility; their
  food security announcements are likely to be received
  with cynicism regardless of content.
- What emotional response would a typical Arab Facebook
  audience have to this post given its content and
  source?
Assign a provisional label:
- SOLIDARITY: content inspires support, pride, or
  empathy (aid delivered, achievement announced,
  community initiative)
- DISMISSAL: content is likely to draw cynicism or
  mockery (empty political promises, overreaching
  official claims, corruption, institutional failure
  in a low-trust context)
- GRIEF: content describes suffering, humanitarian
  crisis, famine, displacement, or injustice
- AMBIVALENT: content is neutral, factual, or
  diplomatic with no clear emotional valence
  (statistical reports, routine meetings, policy
  updates)

STEP 2 - Reaction consistency check (secondary)
Love [x]% | Care [x]% | Wow [x]% | Haha [x]% |
Sad [x]% | Angry [x]%
Are the reactions consistent with your Step 1 label?
- If YES: confirm the label.
- If NO: explain the discrepancy. Common cases:
  * High Love on a suffering post = solidarity with
    victims -> keep SOLIDARITY if content confirms,
    reconsider GRIEF if suffering dominates
  * High Haha on a neutral-seeming post = source
    distrust in a low-credibility context ->
    reconsider DISMISSAL
  * High Sad on a political post = content may be more
    distressing than it appears -> reconsider GRIEF

Your final label should reflect content and source
context, with reactions used to confirm or flag
inconsistencies, not to determine the label
mechanically.
```

## Study B (reaction-blind).

```
SYSTEM: You are classifying Arabic-language Facebook
posts about food insecurity in the MENA region based
on text content only.
Read the post text and consider:
- What is being communicated?
- Who is saying it?
- Is the source credible given the country context?
  Governments in Iraq, Sudan, Syria, Lebanon, Libya,
  and Yemen operate in low-trust institutional
  environments.
Assign one label:
- SOLIDARITY: inspires support, pride, or empathy
- DISMISSAL: likely to draw cynicism or mockery
- GRIEF: describes suffering, crisis, or injustice
- AMBIVALENT: neutral, factual, no clear emotional
```

```
  valence
INPUT: Post text, Page name, Page category,
Country (ISO code)
OUTPUT (JSON only):
{"label": "SOLIDARITY|DISMISSAL|GRIEF|AMBIVALENT"}
```

**Study C (matched ablation).** Both conditions used deterministic decoding (temperature 0, top-$p$ 1) and asked for the label only. The reaction-aware prompt is shown below; the text-only prompt is identical except that it omits the reaction percentages and the references to observed reactions.

| Page | Ctry | Date | Post (Arabic) and English gloss |
|---|---|---|---|
| **Ambivalent** (flat profile; Love 27.9%, Haha 15.5%, Care 15.2%) | | | |
| Sputnik Arabic | RU | Apr 2021 | رئيس الحكومة السورية: روسيا دعمتنا باحتياطيات كافية من القمح<br>*Syrian Prime Minister: Russia has supported us with sufficient wheat reserves.* |
| Al Jazeera Yemen | QA | Oct 2022 | بلغت صادرات أوكرانيا من الحبوب والمواد الغذائية منذ توقيع اتفاق فتح الموانئ ٧١٩ ألفاً و٥٤٩ طناً<br>*Since the port-opening agreement was signed, Ukraine's grain and food exports have reached 719,549 tonnes.* |
| Social Fund for Development (Yemen) | YE | 2021 | نقدم أجور عمل عادلة للعمال نظير إنشائهم خدمات مستدامة تساهم في توفير الأمن الغذائي والمائي<br>*We pay workers fair wages for building sustainable services that contribute to food and water security.* |
| **Dismissal** (Haha 58.0%, Love 15.4%) | | | |
| Maysan News (ميسان نيوز) | IQ | Aug 2022 | قيام الكاظمي بمنح الصلاحيات الكاملة بالتصرف بأموال قانون الأمن الغذائي<br>*Al-Kadhimi grants full powers to dispose of the Food Security Law funds.* |
| Diwan FM | TN | Oct 2022 | قرض جديد لتونس بقيمة ٢٣٣ مليون دولار لتأمين الأمن الغذائي والطاقي<br>*A new $233 million loan for Tunisia to secure food and energy security.* |
| Economy Today (الاقتصاد اليوم) | SY | Jul 2021 | إذا لم تتغير سياساتنا فستحدث كارثة بعجوزات تأمين بعض المحاصيل الأساسية<br>*If our policies do not change, there will be a catastrophe of shortfalls in securing some basic crops.* |
| **Solidarity** (Love 67.7%, Care 12.0%) | | | |
| Information Center, Egypt (مركز المعلومات - مصر) | EG | Apr 2022 | قضية الأمن الغذائي أصبحت قضية استراتيجية ترتبط بالأمن القومي<br>*Food security has become a strategic issue tied to national security.* |
| Agri Eng | EG | 2022 | الصادرات الزراعية تحقق رقماً قياسياً بـ ٦.٥ مليار دولار<br>*Agricultural exports reach a record $6.5 billion.* |
| Social Fund for Development (Yemen) | YE | 2021 | يتحقق الأمن الغذائي عندما تتوافر لجميع الناس الأساسيات الضرورية للحياة<br>*Food security is achieved when all people have the basic necessities of life.* |
| **Grief** (Sad 60.1%, Angry 10.4%) | | | |
| Al Jazeera Yemen | QA | Aug 2022 | أزمة جوع غير مسبوقة تُهدّد اليمن. ١٧.٤ مليون يمني يعانون من انعدام الأمن الغذائي<br>*An unprecedented hunger crisis threatens Yemen; 17.4 million Yemenis suffer from food insecurity.* |
| Al Jazeera Palestine | QA | Oct 2023 | جميع سكان غزة يعانون من مستويات أسوأ من انعدام الأمن الغذائي الحاد<br>*All of Gaza's residents are experiencing worsening levels of acute food insecurity.* |
| Journalist's public page | GB | 2023 | الأمم المتحدة تحذر من أسوأ أزمة غذاء يشهدها السودان<br>*The UN warns of the worst food crisis Sudan has witnessed.* |

**Table A.6:** Representative posts per reaction profile, with English glosses. Ctry is the page-administration country.

```
SYSTEM: You are classifying Arabic-language Facebook
posts about food insecurity in the MENA region. Your
task is to infer the likely audience emotional-response
profile. Use the post text, source context, and the
observed Facebook reaction percentages. Do not infer
only the author's sentiment or writing tone.

Read the post text and consider:
- What is being communicated?
- Who is saying it?
- Is the source credible given the country context?
  Governments in Iraq, Sudan, Syria, Lebanon, Libya,
  and Yemen operate in low-trust institutional
  environments.
- What emotional response would a typical Arab
  Facebook audience have to this post given its
  content, source, and observed reactions?

Assign one label:
- SOLIDARITY: content/reactions indicate support,
  pride, affiliation, or empathy
- DISMISSAL: content/reactions indicate cynicism,
  mockery, scepticism, or ironic rejection
```

```
- GRIEF: content/reactions indicate suffering,
  humanitarian crisis, distress, or injustice
- AMBIVALENT: no clear dominant emotional-response
  profile
INPUT: Post text, Page name, Page category,
Country (ISO code), Reactions: Love [x]% | Care [x]%
| Wow [x]% | Haha [x]% | Sad [x]% | Angry [x]%
OUTPUT: one label
SOLIDARITY | DISMISSAL | GRIEF | AMBIVALENT
```